\documentclass[10pt,twocolumn,letterpaper]{article}
\usepackage{cvpr}
\usepackage{times}
\usepackage{epsfig}
\usepackage{graphicx}
\usepackage{amsmath}
\usepackage{amssymb}
\usepackage{multirow}
\usepackage{float}
\usepackage{tabulary}
\usepackage{tabularx}
\usepackage[tight,normalsize,sf,SF]{subfigure}
\usepackage{color}
\usepackage{soul}
\usepackage{enumitem}
\usepackage{hhline}
\usepackage{algorithm}
\usepackage{algpseudocode}
\usepackage{array}
\usepackage{flushend}
\usepackage{csquotes}
\newcolumntype{P}[1]{>{\centering\arraybackslash}p{#1}}
\DeclareMathSymbol{\Lambda}{\mathalpha}{operators}{3}
\DeclareMathSymbol{\Pi}{\mathalpha}{operators}{5}
\graphicspath{{./figs/}}
\DeclareGraphicsExtensions{.eps,.ps,.pdf,.png,.tiff}

\usepackage[table]{xcolor}

\usepackage[pagebackref=true,breaklinks=true,letterpaper=true,colorlinks,bookmarks=false]{hyperref}

\cvprfinalcopy 

\def\cvprPaperID{3289}

\ifcvprfinal\fi
\begin{document}

\title{Unsupervised Adaptive Re-identification in Open World Dynamic Camera Networks}

\author{Rameswar Panda$^{1,}$\thanks{RP and AB should be considered as joint first authors} \ \ \ \  Amran Bhuiyan$^{2,\ast,}$\thanks{This work was done while AB was a visiting student at UC Riverside}  \ \ \ \ Vittorio Murino$^{2}$ \ \ \ \ Amit K. Roy-Chowdhury$^{1}$\\
$^{1}$ Department of ECE \ \ \ \ \ \ \ \  $^{2}$Pattern Analysis and Computer Vision (PAVIS)  \\
UC Riverside \ \ \ \ \ \ \ \ \ \ \ \ \ \ \ \ \ \ \ \ \ \ \ \ \ \ \ Istituto Italiano di Tecnologia, Italy\\
{\tt\small \{rpand002@,amitrc@ece.\}ucr.edu \ \ \ \{amran.bhuiyan,vittorio.murino\}@iit.it}
}

\maketitle

\begin{abstract} 
	
Person re-identification is an open and challenging problem in computer vision. Existing approaches have concentrated on either designing the best feature representation or learning optimal matching metrics in a static setting where the number of cameras are fixed in a network. Most approaches have neglected the dynamic and open world nature of the re-identification problem, where a new camera may be temporarily inserted into an existing system to get additional information. To address such a novel and very practical problem, we propose an unsupervised adaptation scheme for re-identification models in a dynamic camera network. First, we formulate a domain perceptive re-identification method based on geodesic flow kernel that can effectively find the best source camera (already installed) to adapt with a newly introduced target camera, without requiring a very expensive training phase. Second, we introduce a transitive inference algorithm for re-identification that can exploit the information from best source camera to improve the accuracy across other camera pairs in a network of multiple cameras. Extensive experiments on four benchmark datasets demonstrate that the proposed approach significantly outperforms the state-of-the-art unsupervised learning based alternatives whilst being extremely efficient to compute.       
 
\end{abstract}
\vspace{-4mm}

\section{Introduction}
\label{sec:intro}

Person re-identification (re-id), which addresses the problem of matching people across non-overlapping views in a multi-camera system, has drawn a great deal of attention in the last few years~\cite{karanam2016comprehensive,vezzani2013people,zheng2016person}. 
Much progress has been made in developing methods that seek either the best feature representations (e.g.,~\cite{ziyanpami,lisanti2015person,bazzani2013symmetry,liu2012person}) or propose to learn optimal matching metrics (e.g.,~\cite{liao2015person,paisitkriangkrai2015learning,liao2015person,xiong2014person}). While they have obtained reasonable performance on commonly used benchmark datasets (e.g.,~\cite{6239203,garcia2015person,das2014consistent,zheng2015scalable}), we believe that these approaches have not yet considered a fundamental related problem: \textit{Given a camera network where the inter-camera transformations/distance metrics have been learned in an intensive training phase, how can we incorporate a new camera into the installed system with minimal additional effort?} This is an important problem to address in realistic open-world re-identification scenarios, where a new camera may be temporarily inserted into an existing system to get additional information. 

\begin{figure}
	\centering
	\begin{tabular}{c}
		\includegraphics[scale=0.11]{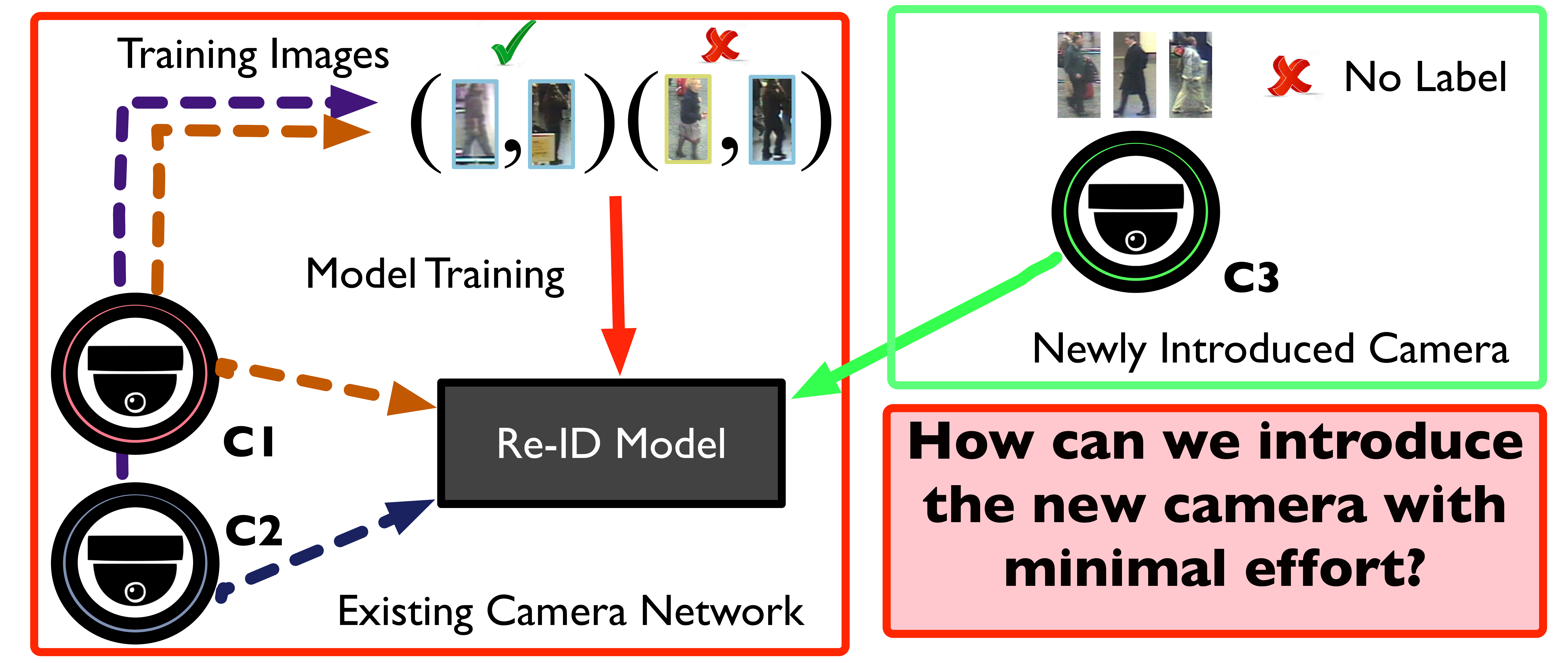}
	\end{tabular}
	\vspace{1mm}
	\caption
	{Consider an existing network with two cameras $\textbf{\texttt{C}}_1$ and $\textbf{\texttt{C}}_2$ where we have learned a re-id model using pair-wise training data from both of the cameras. During the operational phase, a new camera $\textbf{\texttt{C}}_3$ is introduced to cover a certain area that is not well covered by the existing 2 cameras. Most of the existing methods do not consider such dynamic nature of a re-id model. In contrast, we propose to adapt the existing re-id model in an unsupervised way: \textit{what is the best source camera to pair with the new camera and how can we exploit the best source camera to improve the matching accuracy across the other cameras.}}
	\label{fig:MotivationFigure}
	\vspace{-5mm}
\end{figure} 

To illustrate such a problem, let us consider a scenario with $\mathcal{N}$ cameras for which we have learned the optimal pair-wise distance metrics, so providing high re-id accuracy for all camera pairs. However, during a particular event, a new camera may be temporarily introduced to cover a certain related area that is not well-covered by the existing network of $\mathcal{N}$ cameras (See Fig.~\ref{fig:MotivationFigure} for an example). Despite the dynamic and open nature of the world, almost all work in re-identification assumes a \textit{static} and \textit{closed} world model of the re-id problem where the number of cameras is fixed in a network. Given a newly introduced camera, traditional re-id methods will try to relearn the inter-camera transformations/distance metrics using a costly training phase.
This is impractical since labeling data in the new camera and then learning transformations with the others is time-consuming, and defeats the entire purpose of temporarily introducing the additional camera.
Thus, there is a pressing need to develop \textit{unsupervised} learning models for re-identification that can work in such dynamic camera networks.

Domain adaptation~\cite{daume2009frustratingly,kulis2011you} has recently been successful in many classical vision problems such as object recognition~\cite{jie2011multiclass,saenko2010adapting,gopalan2011domain} and activity classification~\cite{ma2014harnessing,yang2013related} with multiple classes or domains. The main objective is to scale learned systems from a source domain to a target domain without requiring prohibitive amount of training data in the target domain.
Considering a newly introduced camera as target domain, we pose an important question in this paper: \textit{Can unsupervised domain adaptation be leveraged upon for re-identification in a dynamic camera network?}

Unlike classical vision problems, e.g., object recognition~\cite{saenko2010adapting}, domain adaptation for re-id has additional challenges. A central issue in domain adaptation is that from \textit{which source to transfer}. When there is only one source of information available which is highly relevant to the task of interest, then domain adaptation is much simpler than in the more general and realistic case where there are multiple sources of information of greatly varying relevance. Re-identification in a dynamic network falls into the latter, more difficult, case. Specifically, given multiple source cameras (already installed) and a target camera (newly introduced), \textit{how can we select the best source camera to pair with the target camera?} The problem can be easily extended to multiple additional cameras being introduced.

Moreover, once the best source camera is identified, \textit{how can we exploit this information to improve the re-identification accuracy of other camera pairs?} For instance, let us consider $\textbf{\texttt{C}}_1$ being the best source camera for the newly introduced camera $\textbf{\texttt{C}}_3$ in Fig.~\ref{fig:MotivationFigure}. Once the pair-wise distance metric between $\textbf{\texttt{C}}_1$ and $\textbf{\texttt{C}}_3$ is obtained, can we exploit this information to improve the re-id accuracy across ($\textbf{\texttt{C}}_2$--$\textbf{\texttt{C}}_3$)? This is an especially important problem because it will allow us to now match data in the newly inserted target camera $\textbf{\texttt{C}}_3$ with all the previously installed cameras. 
\vspace{-1mm}
\subsection{Overview of Solution Strategy}
\vspace{-1mm}
To address the first challenge, we propose an unsupervised re-identification method based on geodesic flow kernel~\cite{gong2012geodesic,gopalan2011domain} that can effectively find the best source camera to adapt with a target camera. Given camera pairs, each consisting of 1 (out of $\mathcal{N}$) source camera and a target camera, we first compute a kernel over the subspaces representing the data of both cameras and then use it to find the kernel distance across the source and target camera. Then, we rank the source cameras based on the average distance and choose the one with lowest distance as the best source camera to pair with the target camera. This is intuitive since a camera which is closest to the newly introduced camera will give the best re-id performance on the target camera and hence, is more likely to adapt better than others. In other words, a source camera with lowest distance with respect to a target camera indicates that both of the sensors could be similar to each other and their features may be similarly distributed. Note that we learn the kernel with the labeled data from the source camera only.

To address the second challenge, we introduce a transitive inference algorithm for re-identification that can exploit information from best source camera to improve accuracy across other camera pairs. 
With regard to the example in Fig.~\ref{fig:MotivationFigure} in which source camera $\textbf{\texttt{C}}_1$ best matches with target camera $\textbf{\texttt{C}}_3$, our proposed transitive algorithm establishes a path between camera pair ($\textbf{\texttt{C}}_2$ -- $\textbf{\texttt{C}}_3$) by marginalization over the domain of possible appearances in best source camera $\textbf{\texttt{C}}_1$. Specifically, $\textbf{\texttt{C}}_1$ plays the role of a \enquote{connector} between $\textbf{\texttt{C}}_2$ and $\textbf{\texttt{C}}_3$. Experiments show that this approach indeed increases the overall re-id accuracy in a network by improving matching performance across camera pairs, while exploiting side information from best source camera.
\vspace{-0.5mm}
\subsection{Contributions}
\vspace{-1mm}
We address a novel, and very practical, problem in this paper --- how to add one or more cameras temporarily to an existing network and exploit it for re-identification, without also adding a very expensive training phase. To the best of our knowledge, this is the first time such a problem is being addressed in re-identification research. Towards solving this problem, we make the following contributions: 
\newline
{(i)} An unsupervised re-identification approach based on geodesic flow kernel that can find the best source camera to adapt with a newly introduced target camera in a dynamic camera network; {(ii)} a transitive inference algorithm to exploit the side information from the best source camera to improve the matching accuracy across other camera pairs; {(iii)} rigorous experiments validating the advantages of our approach over existing alternatives on multiple benchmark datasets with variable number of cameras.                              
\vspace{-2mm}
\section{Related Work}
\label{sec:related work}
\vspace{-1mm}
Person re-identification has been studied from different perspectives (see~\cite{zheng2016person} for a recent survey). Here, we focus on some representative methods closely related to our work.    

\vspace{1mm}
\textbf{Supervised Re-identification.} Most existing person re-identification techniques are based on supervised learning. These methods either seek the best feature representation~\cite{ziyanpami,lisanti2015person,bazzani2013symmetry,martinel2015re,chakraborty2016network,roy2012camera,bhuiyan2015exploiting,bhuiyan2014person} or learn discriminant metrics/dictionaries~\cite{liao2015person,zheng2013reidentification,pedagadi2013local,li2013learning,liao2015efficient,tao2013person,hirzer2012relaxed,karanam2015person,zhao2014learning} that yield an optimal matching score between two cameras or between a gallery and a probe image. Recently, deep learning methods have been applied to person re-identification~\cite{yi2014deep,yan2016person,xiao2016learning,cheng2016person,wu2016personnet,liu2016multi}. Combining feature representation and metric learning with end-to-end deep neural networks is also a recent trend in re-identification~\cite{ahmed2015improved,li2014deepreid,xiao2016end}. Considering that a modest-sized camera network can easily have hundreds of cameras, these supervised re-id models will require huge amount of labeled data which is difficult to collect in real-world settings. In an effort to bypass tedious labeling of training data in supervised re-id models, there has been recent interest in using active learning for re-identification that intelligently selects unlabeled examples for the experts to label in an interactive manner~\cite{liu2013pop,das2015active,martinel2016temporal,wang2016human,das2017continuous}. However, all these approaches consider a static camera network unlike the problem domain we consider.

\vspace{1mm}
\textbf{Unsupervised Re-identification.} Unsupervised learning models have received little attention in re-identification because of their weak performance on benchmarking datasets compared to supervised methods. Representative methods along this direction use either hand-crafted appearance features~\cite{ma2012local,liu2014fly,ma2012bicov,cheng2011custom} or saliency statistics~\cite{zhao2013unsupervised} for matching persons without requiring huge amount of labeled data. Recently, dictionary learning based methods have also been utilized for re-identification in an unsupervised setting~\cite{kodirov2016person,liu2014semi,kodirov2015dictionary}. Although being scalable in real-world settings, these approaches have not yet considered the dynamic nature of the re-id problem, where new cameras can be introduced at any time to an existing network.

\vspace{1mm}
\textbf{Open World Re-Identification.} Open world recognition has been introduced in~\cite{bendale2015towards} as an attempt to move beyond the dominant static setting to a dynamic and open setting where the number of training images and classes are not fixed in recognition. Inspired by such approaches, recently there have been few works in re-identification~\cite{zheng2016towards,campslab} which try to address the open world scenario by assuming that gallery and probe sets contain different identities of persons.
Unlike such approaches, we consider yet another important aspect of open world re-identification where the camera network is dynamic and the system has to incorporate a new camera with minimal additional effort.

\vspace{1mm}
\textbf{Domain Adaptation.} Domain adaptation, which aims to adapt a source domain to a target domain, has been successfully used in many areas of computer vision and machine learning. Despite its applicability in classical vision tasks, domain adaptation for re-identification still remains as a challenging and under addressed problem. Only very recently, domain adaptation for re-id has begun to be considered~\cite{li2012human,zheng2012transfer,wang2015cross,ma2015cross}. However, these studies consider only improving the re-id performance in a static camera network with fixed number of cameras.    

\vspace{1mm}
To the best of our knowledge, this is the first work to address the intrinsically dynamic nature of re-identification in unconstrained open world settings, i.e., scenarios where new camera(s) can be introduced to an existing network, and which, the system will have to incorporate for re-identification with minimal to no human supervision.
\vspace{-2.5mm}
\section{Methodology}
\label{sec:Methodology}

\vspace{-1mm} 
To adapt re-id models in a dynamic camera network, we first formulate a domain perceptive re-identification approach based on geodesic flow kernel which can effectively find the best source camera (out of multiple installed ones) to pair with a newly introduced target camera with minimal additional effort (Section~\ref{sec:Best Source}). Then, to exploit information from the best source camera, we propose a transitive inference algorithm that improves the matching performance across other camera pairs in a network (Section~\ref{sec:Transitive}). 
\vspace{-1mm}
\subsection{Initial Setup}
\label{sec:Preliminaries}
\vspace{-1mm}
Our proposed framework starts with an installed camera network where the discriminative distance metrics between each camera pairs is learned using a off-line intensive training phase. Let there be $\mathcal{N}$ cameras in a network and the number of possible camera pairs is $\binom{\mathcal{N}}{2}$.
Let $\{(\mathbf{x}_i^{\mathcal{A}},\mathbf{x}_i^{\mathcal{B}})\}_{i=1}^m$ be a set of training samples, where $\mathbf{x}_i^{\mathcal{A}} \in \mathbb{R}^{{D}}$ represents feature representation of training a sample from camera view ${\mathcal{A}}$ and $\mathbf{x}_i^{\mathcal{B}} \in \mathbb{R}^{D}$ represents feature representation of the same person in a different camera view ${\mathcal{B}}$. 
We assume that the provided training data is for the task of single-shot person re-identification, i.e., there exists only two images of the same person -- one image taken from camera view ${\mathcal{A}}$ and another image taken from camera view ${\mathcal{B}}$.

Given the training data, we follow KISS metric learning (KISSME)~\cite{kostinger2012large} and compute the pairwise distance matrices such that distance between images of the same individual is less than distance between images of different individuals.
The basic idea of KISSME is to learn the Mahalanobis distance by considering a log likelihood
ratio test of two Gaussian distributions. The likelihood ratio
test between dissimilar pairs and similar pairs can be
written as
\vspace{-1mm}
\begin{equation}
	\begin{gathered}
		\label{eq:kissme1} 
		\mathcal{R}(\mathbf{x}_i^{\mathcal{A}},\mathbf{x}_j^{\mathcal{B}})=\text{log}\dfrac{\frac{1}{\mathcal{C}_\mathcal{D}}\text{exp}(-\frac{1}{2}\mathbf{x}_{ij}^T\Sigma_\mathcal{D}^{-1}\mathbf{x}_{ij})}{\frac{1}{\mathcal{C}_\mathcal{S}}\text{exp}(-\frac{1}{2}\mathbf{x}_{ij}^T\Sigma_\mathcal{S}^{-1}\mathbf{x}_{ij})}
	\end{gathered} 
		\vspace{-1mm}
\end{equation}
where $\mathbf{x}_{ij} = \mathbf{x}_i^{\mathcal{A}}-\mathbf{x}_j^{\mathcal{B}}$, $\mathcal{C}_\mathcal{D}=\sqrt{2\pi|\Sigma_\mathcal{D}|}$, $\mathcal{C}_\mathcal{S}=\sqrt{2\pi|\Sigma_\mathcal{S}|}$, $\Sigma_\mathcal{D}$ and $\Sigma_\mathcal{S}$ are covariance matrices of dissimilar and similar pairs respectively. With simple manipulations, (\ref{eq:kissme1}) can be written as
\vspace{-1mm}
\begin{equation}
	\small
\begin{gathered}
\label{eq:kissme2} 
\mathcal{R}(\mathbf{x}_i^{\mathcal{A}},\mathbf{x}_j^{\mathcal{B}})=\mathbf{x}_{ij}^T\mathbf{M}\mathbf{x}_{ij} 
\end{gathered}
\end{equation}
where $\mathbf{M}=\Sigma_\mathcal{S}^{-1}-\Sigma_\mathcal{D}^{-1}$ is the Mahalanobis distance between covariances associated to a pair of cameras. We follow~\cite{kostinger2012large} and clip the spectrum by an eigen-analysis to ensure $\mathbf{M}$ is positive semi-definite.

Note that our approach is agnostic to the choice of metric learning algorithm used to learn the optimal metrics across camera pairs in an already installed network. We adopt KISSME in this work since it is simple to compute and has shown to perform satisfactorily on the person re-id problem. We will also show an experiment using Logistic Discriminant-based Metric Learning (LDML)~\cite{guillaumin2009you} instead of KISSME later in Section~\ref{sec:lfda}. 
\vspace{-1mm}
\subsection{Discovering the Best Source Camera}
\label{sec:Best Source}
\textbf{Objective.} Given an existing camera network where optimal matching metrics across all possible camera pairs are computed using the above training phase, our first objective is to select the best source camera which has the lowest kernel distance with respect to the newly inserted camera. Labeling data in the new camera and then learning distance metrics with the already existing $\mathcal{N}$ cameras is practically impossible since labeling all the samples may often require tedious human labor. To overcome such an important problem, we adopt an unsupervised strategy based on geodesic flow kernel~\cite{gong2012geodesic,gopalan2011domain} to learn the metrics without requiring any labeled data from the newly introduced camera.

\vspace{1mm}
\textbf{Approach Details.} Our approach for discovering the best source camera consists of the following steps: (i) compute geodesic flow kernels between the new (target) camera and other existing cameras (source); (ii) use the kernels to determine the distance between them; (iii) rank the source cameras based on distance with respect to the target camera and choose the one with the lowest as best source camera.     

Let $\{\mathcal{X}^s\}_{s=1}^{\mathcal{N}}$ be the $\mathcal{N}$ source cameras and $\mathcal{X}^{\mathcal{T}}$ be the newly introduced target camera. To compute the kernels in an unsupervised way, we extend a previous method~\cite{gong2012geodesic} that adapts classifiers in the context of object recognition to re-identification in a dynamic camera network. The main idea of our approach is to compute the low-dimensional subspaces representing data of two cameras (one source and one target) and then map them to two points on a Grassmannian\footnote{Let $d$ being the dimension of the subspace, the collection of all $d$-dimensional subspaces form the Grasssmannian.}. Intuitively, if these two points are close on the Grassmannian, then the computed kernel would provide high matching performance on the target camera. In other words, both of the cameras could be similar to each other and their features may be similarly distributed over the corresponding subspaces. For simplicity, let us assume we are interested in computing the kernel matrix $\mathbf{K}^{\mathcal{ST}}\in \mathbb{R}^{D \times D}$ between the source camera $\mathcal{X}^{\mathcal{S}}$ and a newly introduced target camera $\mathcal{X}^{\mathcal{T}}$. Let  $\mathcal{\tilde{X}}^{\mathcal{S}} \in \mathbb{R}^{D \times d}$ and $\mathcal{\tilde{X}}^{\mathcal{T}} \in \mathbb{R}^{D \times d}$ denote the $d$-dimensional subspaces, computed using Partial Least Squares (PLS) and Principal Component Analysis (PCA) on the source and target camera, respectively. Note that we cannot use PLS on the target camera since it is a supervised dimension reduction technique and requires label information for computing the subspaces.  

Given both of the subspaces, the closed loop solution to the geodesic flow kernel between the source and target camera is defined as
\vspace{-2mm}
\begin{equation}
\begin{gathered}
\label{eqn:gfk1}
{\mathbf{x}_i^{\mathcal{S}}}^T \mathbf{K}^{\mathcal{ST}} \mathbf{x}_j^{\mathcal{T}} = \int_0^1 (\psi(\mathbf{y})^T \mathbf{x}_i^{\mathcal{S}})^T (\psi(\mathbf{y}) \mathbf{x}_j^{\mathcal{T}})\ d\textbf{y} 
\end{gathered}
\vspace{-2mm}
\end{equation}
where $\mathbf{x}_i^{\mathcal{S}}$ and $\mathbf{x}_j^{\mathcal{T}}$ represent feature descriptor of $i$-th and $j$-th sample in source and target camera respectively. $\psi(\mathbf{y})$ is the geodesic flow parameterized by a continuous variable $\textbf{y}\in [0,1]$ and represents how to smoothly project a sample from the original $D$-dimensional feature space onto the corresponding low dimensional subspace. The  geodesic flow $\psi(\mathbf{y})$ over two cameras can be defined as~\cite{gong2012geodesic}, 
\vspace{-1mm}
\begin{equation}
\begin{gathered}
\label{eqn:gfk2}
\psi(\mathbf{y}) =
\begin{cases}
\mathcal{\tilde{X}}^{\mathcal{S}} &  \text{if } \mathbf{y}=0\\
\mathcal{\tilde{X}}^{\mathcal{T}}  &  \text{if } \mathbf{y}=1\\
\mathcal{\tilde{X}}^{\mathcal{S}} \mathcal{U}_1\mathcal{V}_1(\mathbf{y})- \mathcal{\tilde{X}}^{\mathcal{S}}_o\mathcal{U}_2\mathcal{V}_2(\mathbf{y}) &  \text{otherwise}\\
\end{cases}
\end{gathered} 
\vspace{-1mm}
\end{equation} 
where $\mathcal{\tilde{X}}^{\mathcal{S}}_o \in \mathbb{R}^{D \times (D-d)}$ is the orthogonal matrix to $\mathcal{\tilde{X}}^{\mathcal{S}}$ and $\mathcal{U}_1,\mathcal{V}_1,\mathcal{U}_2,\mathcal{V}_2$ are given by the following pairs of SVDs,
\vspace{-1mm}
\begin{equation}
\begin{gathered}
\label{eqn:gfk3}
{\mathcal{X}^{\mathcal{S}}}^T \mathcal{X}^{\mathcal{T}}=\mathcal{U}_1\mathcal{V}_1\mathcal{P}^T, \ {\mathcal{X}^{\mathcal{S}}_o}^T \mathcal{X}^{\mathcal{T}}=-\mathcal{U}_2\mathcal{V}_2\mathcal{P}^T.
\end{gathered} 
\vspace{-1mm}
\end{equation} 
With the above defined matrices, $\mathbf{K}^{\mathcal{ST}}$ can be computed as 
\vspace{-1mm}
\begin{equation}
\label{eqn:gfk4}
\mathbf{K}^{\mathcal{ST}} = 
\begin{bmatrix}
\mathcal{\tilde{X}}^{\mathcal{S}} \mathcal{U}_1 & \mathcal{\tilde{X}}^{\mathcal{S}}_o\mathcal{U}_2 \\
\end{bmatrix}
\mathcal{G}
\begin{bmatrix}
\mathcal{U}_1^T{\mathcal{X}^{\mathcal{S}}}^T  \\
\mathcal{U}_2^T {\mathcal{X}^{\mathcal{S}}_o}^T \\
\end{bmatrix}
\vspace{-1mm}
\end{equation}
where $\mathcal{G}= \begin{bmatrix}
	\text{diag}[1+\frac{\text{sin}(2\theta_i)}{2\theta_i}] & \text{diag}[\frac{(\text{cos}(2\theta_i)-1)}{2\theta_i}] \\
	\text{diag}[\frac{(\text{cos}(2\theta_i)-1)}{2\theta_i}] & \text{diag}[1-\frac{\text{sin}(2\theta_i)}{2\theta_i}] \\
\end{bmatrix}$ and $[\theta_i]_{i=1}^d$ represents the principal angles between source and target camera. 
Once we compute all pairwise geodesic flow kernels between a target camera and source cameras using (\ref{eqn:gfk4}), our next objective is to find the distance across all those pairs. 
A source camera which is closest to the newly introduced camera is more likely to adapt better than others. We follow~\cite{phillips2011gentle} to compute distance between a target camera and a source camera pair. Specifically, given a kernel matrix $\mathbf{K}^{\mathcal{ST}}$, the distance between data points of a source and target camera is defined as
\vspace{-1mm}
\begin{equation}
\small
\begin{gathered}
\label{eqn:gfk5}
\mathbf{D}^{\mathcal{ST}}(\mathbf{x}_i^\mathcal{S},\mathbf{x}_j^\mathcal{T})={\mathbf{x}_i^\mathcal{S}}^T\mathbf{K}^{\mathcal{ST}}{\mathbf{x}_i^\mathcal{S}}+{\mathbf{x}_j^\mathcal{T}}^T\mathbf{K}^{\mathcal{ST}}{\mathbf{x}_j^\mathcal{T}}-2{\mathbf{x}_i^\mathcal{S}}^T\mathbf{K}^{\mathcal{ST}}{\mathbf{x}_j^\mathcal{T}}
\end{gathered} 
\end{equation} 
where $\mathbf{D}^{\mathcal{ST}}$ represents the kernel distance matrix defined over a source and target camera.
We compute the average of a distance matrix $\mathbf{D}^{\mathcal{ST}}$ and consider it as the distance between two camera pairs.  
Finally, we chose the one that has the lowest distance as the best source camera to pair with 
the newly introduced target camera. 

\textbf{Remark 1.} Note that we do not use any labeled data from the target camera to either compute the geodesic flow kernels in (\ref{eqn:gfk4}) or the kernel distance matrices in (\ref{eqn:gfk5}). Hence, our approach can be applied to adapt re-id models in a large-scale camera network with minimal additional effort.  

\textbf{Remark 2.} We assume that the newly introduced camera will be close to at least one of the installed ones since we consider them to be operating in the same time window and thus have similar environmental factors. Moreover, our approach is not limited to a single camera and can be easily extended to even more realistic scenarios where multiple cameras are introduced to an existing network at the same time. One can easily find a common best source camera based on lowest average distance to pair with all the new cameras, or multiple best source cameras, one for each target camera, in an unsupervised way similar to the above approach. Experiments on a large-scale network of 16 cameras show the effectiveness of our approach in scenarios where multiple cameras are introduced at the same time (See Section ~\ref{sec:multi}).    

\subsection{Transitive Inference for Re-identification}
\label{sec:Transitive}
\vspace{1mm}
\textbf{Objective.} In the previous section we have presented a domain perceptive re-identification approach that can effectively find a best source camera to pair with the target camera in a dynamic camera network. Once the best source camera is identified, the next question is: \textit{can we exploit the best source camera information to improve the re-identification accuracy of other camera pairs?} 

\vspace{1mm}
\textbf{Approach Details.} Let $\{\textbf{M}^{ij}\}_{i,j=1,i<j}^{\mathcal{N}}$ be the optimal pair-wise metrics learned in a network of $\mathcal{N}$ cameras following Section~\ref{sec:Preliminaries} and $\mathcal{S}^{\star}$ be the best source camera for a newly introduced target camera $\mathcal{T}$ following Section~\ref{sec:Best Source}.

Motivated by the effectiveness of Schur product in operations research~\cite{kou2014enhancing1}, we develop a simple yet effective transitive algorithm for exploiting information from the best source camera. Schur product (a.k.a. Hadamard product) has been an important tool for improving the matrix consistency and reliability in multi-criteria decision making. Our problem naturally fits to such decision making systems since our goal is to establish a path between two cameras via the best source camera. Given the best source camera $\mathcal{S}^{\star}$, we compute the kernel matrix between remaining source cameras and the target camera as follows,
\vspace{-1.5mm}
\begin{equation}
\begin{gathered}
\label{eqn:trans1}
{\mathbf{\tilde K}^{\mathcal{ST}}} = {\mathbf{M}^{\mathcal{S}\mathcal{S}^{\star}}} \odot {\mathbf{K}^{\mathcal{S}^{\star}\mathcal{T}}}, \ \forall [\mathcal{S}]_{i=1}^{\mathcal{N}}, \ \ \mathcal{S}\neq\mathcal{S}^{\star}
\end{gathered} 
\vspace{-1.5mm}
\end{equation}
where ${\mathbf{\tilde K}^{\mathcal{ST}}}$ represents the updated kernel matrix between source camera $\mathcal{S}$ and target camera $\mathcal{T}$ by exploiting information from best source camera $\mathcal{S}^{\star}$. The operator $\odot$ denotes Schur product of two matrices. Eq.~\ref{eqn:trans1} establishes an indirect path between camera pair ($\mathcal{S}$,$\mathcal{T}$) by marginalization over the domain of possible appearances in best source camera $\mathcal{S}^{\star}$. In other words, camera $\mathcal{S}^{\star}$ plays a role of connector between the target camera $\mathcal{T}$ and all other source cameras. 

Summarizing, to adapt re-id models in a dynamic network, we use the kernel matrix ${\mathbf{K}^{\mathcal{S}^{\star}\mathcal{T}}}$ computed using (\ref{eqn:gfk4}) to obtain the re-id accuracy across the newly inserted target camera and best source camera, whereas we use the updated kernel matrices, computed using (\ref{eqn:trans1}) to find the matching accuracy across the target camera and remaining source cameras in an existing network.

\textbf{{Remark 3.}} While there are more sophisticated strategies to incorporate the side information, the reason to adopt a simple weighting approach as in (\ref{eqn:trans1}) is that by doing so we can scale the re-identification models easily to a large scale network involving hundreds of cameras in real-time applications. Furthermore, the proposed transitive algorithm performs satisfactorily in several dynamic camera networks as illustrated in Section~\ref{sec:Experiments}.    
\vspace{-1mm}
\subsection{Extension to Semi-supervised Adaptation}
\label{sec:semi}
Although our framework is designed for unsupervised adaptation of re-id models, it can be easily extended if labeled data from the newly introduced target camera becomes available. Specifically, the label information from target camera can be encoded while computing subspaces. That is, instead of using PCA for estimating the subspaces, we can use Partial Least Squares (PLS) to compute the discriminative subspaces on the target data by exploiting the labeled information. PLS has shown to be effective in finding discriminative subspaces by projecting data with labeled information to a common subspace~\cite{geladi1986partial,schwartz2009human}. This essentially leads to semi-supervised adaptation of re-id models in a dynamic camera network (See experiments in Sec~\ref{sec:semiexp}).

\vspace{-1mm}
\section{Experiments}
\label{sec:Experiments}
\vspace{-1mm}
\textbf{Datasets.} We conduct experiments on four different publicly available datasets to verify the effectiveness of our framework, namely 
WARD~\cite{6239203},
 RAiD~\cite{das2014consistent}, 
 SAIVT-SoftBio~\cite{bialkowski2012database}
and Shinpuhkan2014~\cite{kawanishi2014shinpuhkan2014}.
Although there are number of other datasets (\textit{e.g.} ViPeR \cite{gray2008viewpoint}, PRID2011 \cite{roth2014mahalanobis} and CUHK \cite{li2013locally} etc.) for evaluating the performance in re-id, these datasets do not fit our purposes since they have only two cameras or are specifically designed for video-based re-identification~\cite{wang2016person}.
The number of cameras in WARD, RAiD and SAIVT-SoftBio are 3, 4, and 8 respectively. Shinpuhkan2014 is one of the largest publicly available dataset for re-id with 16 cameras.  
Detailed description of these datasets is available in the supplementary material.

\begin{figure*}
	\begin{center}
		\begin{tabular}{ccc}
			\includegraphics[height=.23\linewidth,width=.31\linewidth]{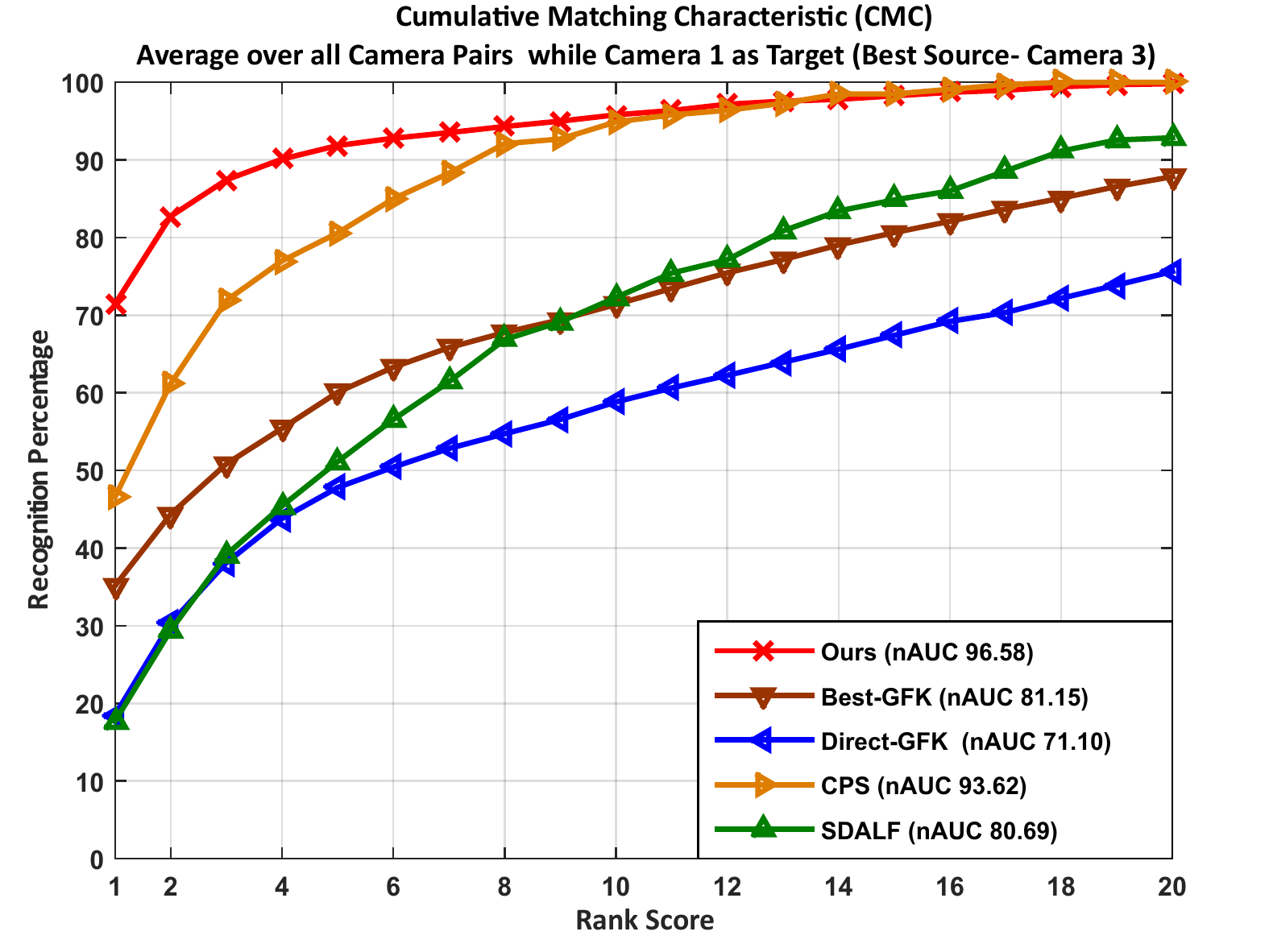} & 
			\includegraphics[height=.23\linewidth,width=.31\linewidth]{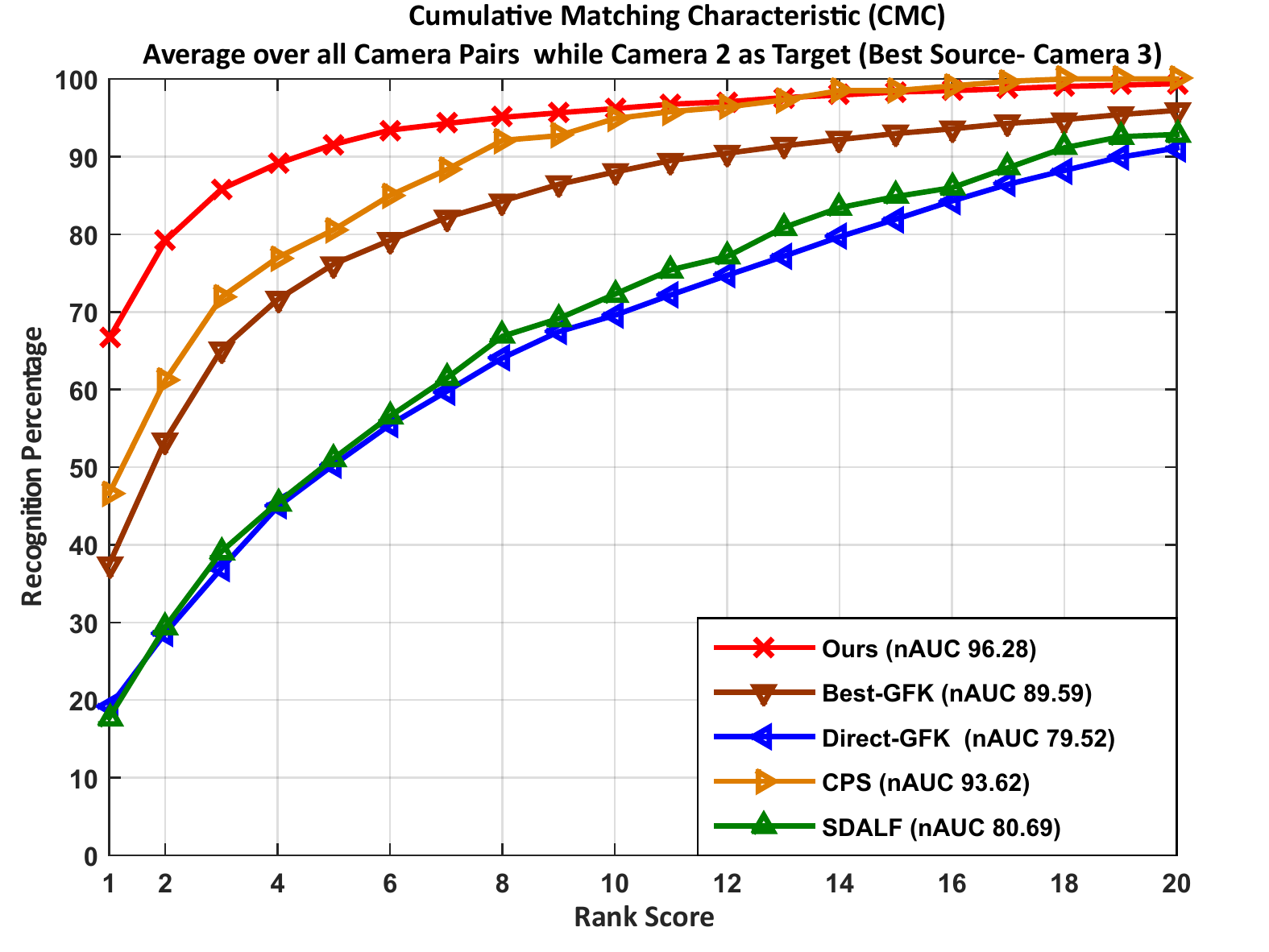}&
			\includegraphics[height=.23\linewidth,width=.31\linewidth]{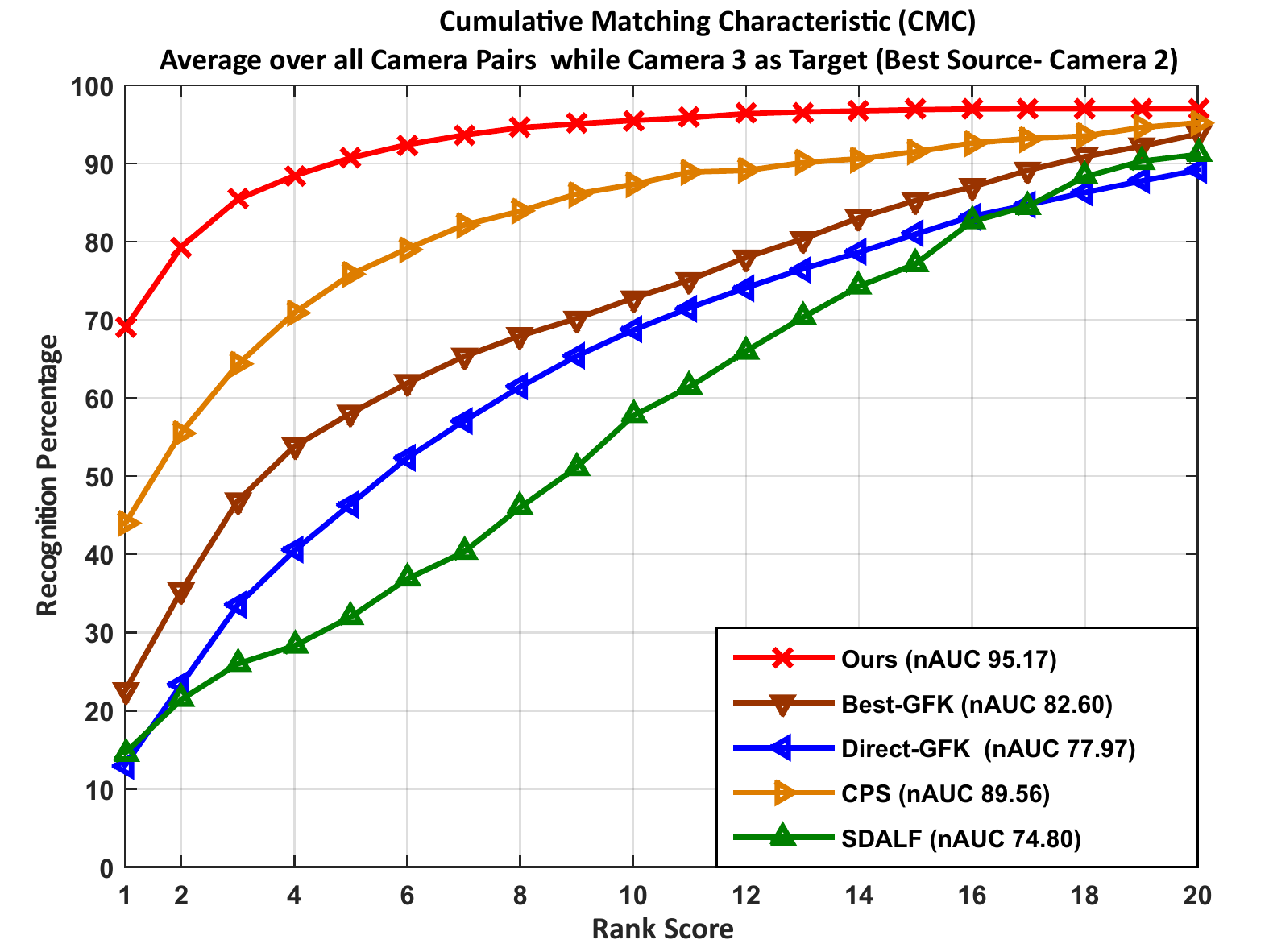}\\
			{\scriptsize (a) Camera 1 as Target } & \scriptsize{ (b) Camera 2 as Target}  & \scriptsize{ (c) Camera 3 as Target}  \\
		\end{tabular}
	\end{center}
	\vspace{-2mm}
	\caption{CMC curves for WARD dataset with 3 cameras. Plots (a, b, c) show the performance of different methods while introducing camera 1, 2 and 3 respectively to a dynamic network. Please see the text in Section~\ref{sec:one} for the analysis of the results. Best viewed in color.}
	\label{fig:WARD}  \vspace{-3mm}
\end{figure*}

\begin{figure}
	\begin{center}
		\begin{tabular}{cc}
			\includegraphics[scale=0.245]{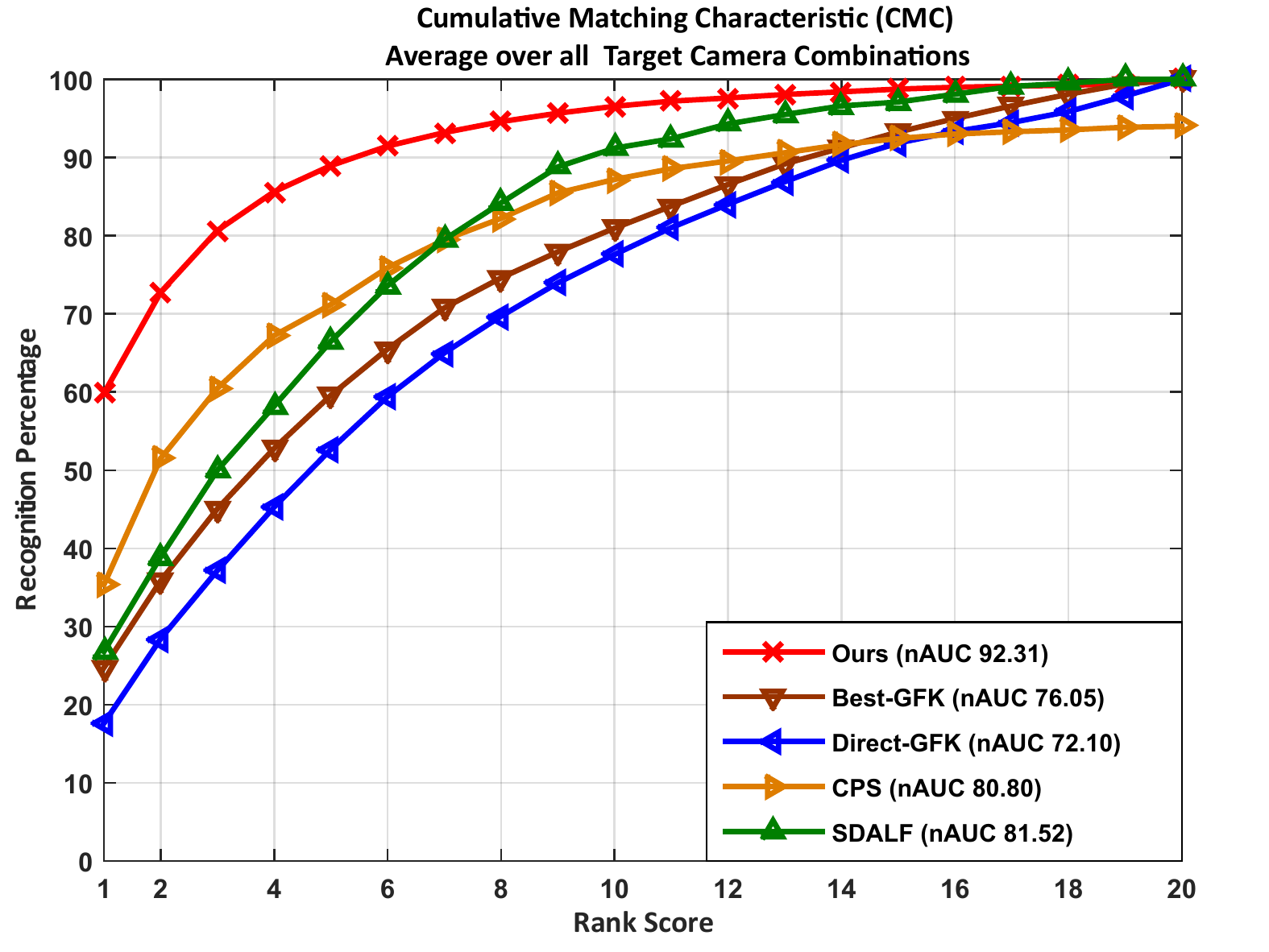} & 
			\includegraphics[scale=0.245]{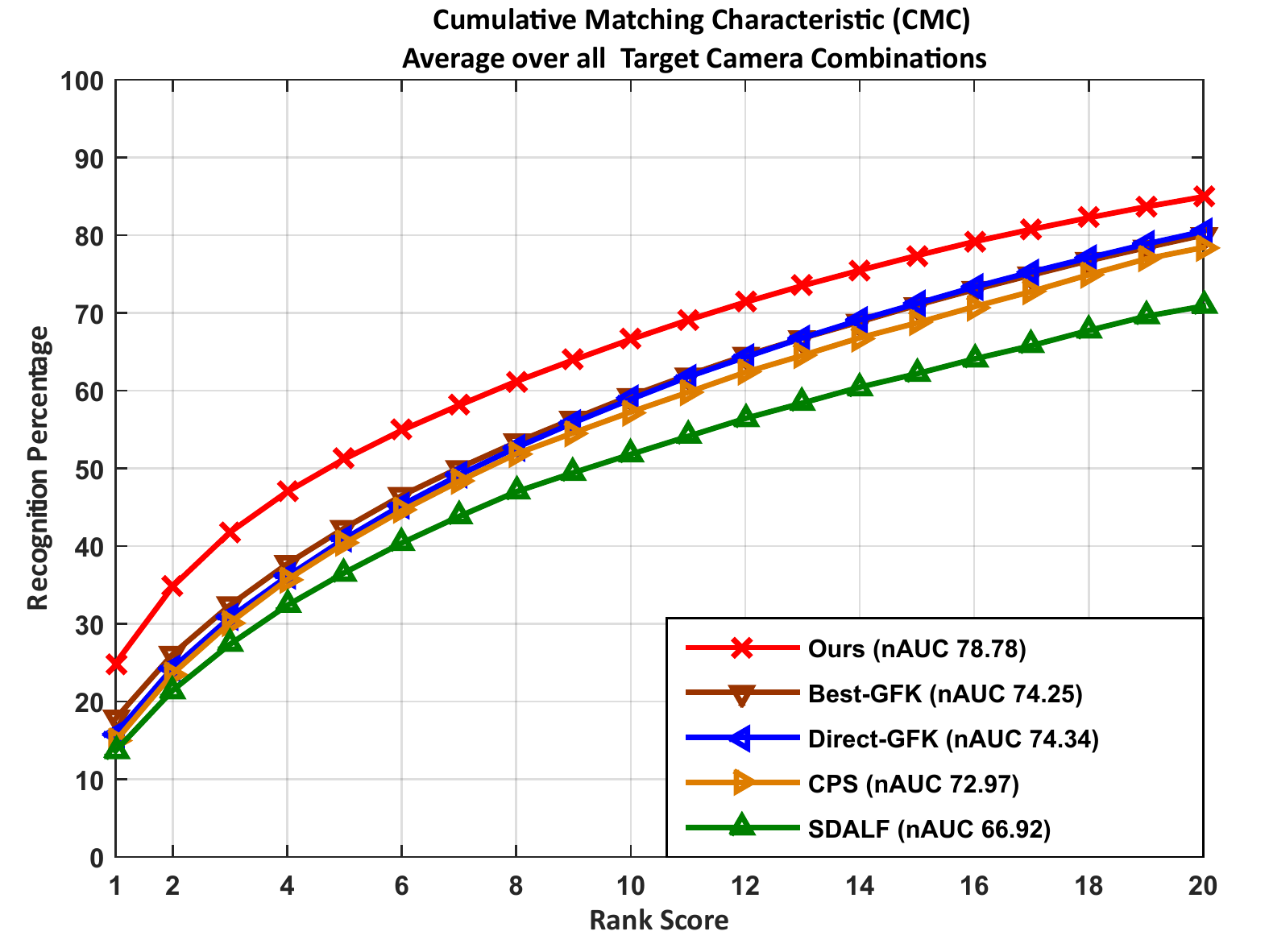}\\
			{\scriptsize (a) RAiD } & \scriptsize{ (b) SAIVT-SoftBio}  \\ 
		\end{tabular}
	\end{center}
	\vspace{-3mm} 
	\caption{CMC curves averaged over all target camera combinations, introduced one at a time. (a) Results on RAiD dataset with 4 cameras (b) Results on SAVIT-SoftBio dataset with 8 cameras. 
	Please see the text in Section~\ref{sec:one} for the analysis of the results.
		}
	\label{fig:RAiD-SAIVT}  \vspace{-4mm} 
\end{figure}
\textbf{Feature Extraction and  Matching.}
The feature extraction stage consists of extracting Local Maximal Occurrence (LOMO) feature proposed in~\cite{lisanti2015person} for person representation. The descriptor has 26,960 dimensions.
We follow~\cite{kostinger2012large,paisitkriangkrai2015learning} and apply principal component analysis to reduce the dimensionality to 100 in all our experiments.
Without low-dimensional feature, it is computationally infeasible to inverse covariance matrices of both similar and dissimilar pairs as discussed in~\cite{kostinger2012large,paisitkriangkrai2015learning}.    
To compute distance between cameras, as well as, re-id matching score, we use kernel distance~\cite{phillips2011gentle} (Eq.~\ref{eqn:gfk5}) for a given projection metric. 

\textbf{Performance Measures.} We show results in terms of recognition rate as Cumulative Matching Characteristic (CMC) curves and normalized Area Under Curve (nAUC) values, as is common practice in re-id literature~\cite{karanam2015person,das2014consistent,martinel2016temporal,zhao2013unsupervised,kodirov2016person}. 
Due to space constraint, we only report average CMC curves for most experiments and leave the full CMC curves in the supplementary material.

\textbf{Experimental Settings.} We maintain following conventions during all our experiments: 
All the images are normalized to 128$\times$64 for being consistent with the evaluations carried out by state-of-the-art methods \cite{bazzani2013symmetry,das2014consistent,cheng2011custom}. Following the literature \cite{das2014consistent,kostinger2012large,lisanti2015person}, the training and testing sets are kept disjoint by picking half of the available data for training set and rest of the half for testing. We repeated each task 10 times by randomly picking 5 images from each identity both for train and test time. The subspace dimension for all the possible combinations are kept at 50. 
\subsection{Re-identification by Introducing a New Camera}
\label{sec:one}
\textbf{Goal.} The goal of this experiment is to analyze the performance of our unsupervised framework while introducing a single camera to an existing network where optimal distance metrics are learned using an intensive training phase. 

\textbf{Compared Methods.} We compare our approach with several unsupervised alternatives which fall into two categories: (i) hand-crafted feature-based methods including \texttt{CPS}~\cite{cheng2011custom} and \texttt{SDALF}~\cite{bazzani2013symmetry}, and (ii) two domain adaptation based methods (\texttt{Best-GFK} and \texttt{Direct-GFK}) based on geodesic flow kernel~\cite{gong2012geodesic}. For \texttt{Best-GFK} baseline, we compute the re-id performance of a camera pair by applying the kernel matrix, ${\mathbf{K}^{\mathcal{S}^{\star}\mathcal{T}}}$ computed between best source and target camera~\cite{gong2012geodesic}, whereas in \texttt{Direct-GFK} baseline, we use the kernel matrix computed directly across source and target camera using (\ref{eqn:gfk4}). 
The purpose of comparing with \texttt{Best-GFK} is to show that the kernel matrix computed across the best source and target camera does not produce optimal re-id performance in computing matching performance across other source cameras and the target camera. 

\vspace{-3mm}

\textbf{Implementation Details.} We use publicly available codes for \texttt{CPS} and \texttt{SDALF} and tested on our experimented datasets. 
We also implement both \texttt{Best-GFK} and \texttt{Direct-GFK} baselines under the same experimental settings as mentioned earlier to have a fair comparison with our proposed method. For all the datasets, we considered one camera as newly introduced target camera and all the other as source cameras. We considered all the possible combinations for conducting experiments. We first pick which source camera matches best with the target one, and then, use the proposed transitive algorithm to compute the re-id performance across remaining camera pairs.  
\begin{figure*}
	\begin{center}
		\begin{tabular}{ccc}
			\includegraphics[height=.23\linewidth,width=.31\linewidth]{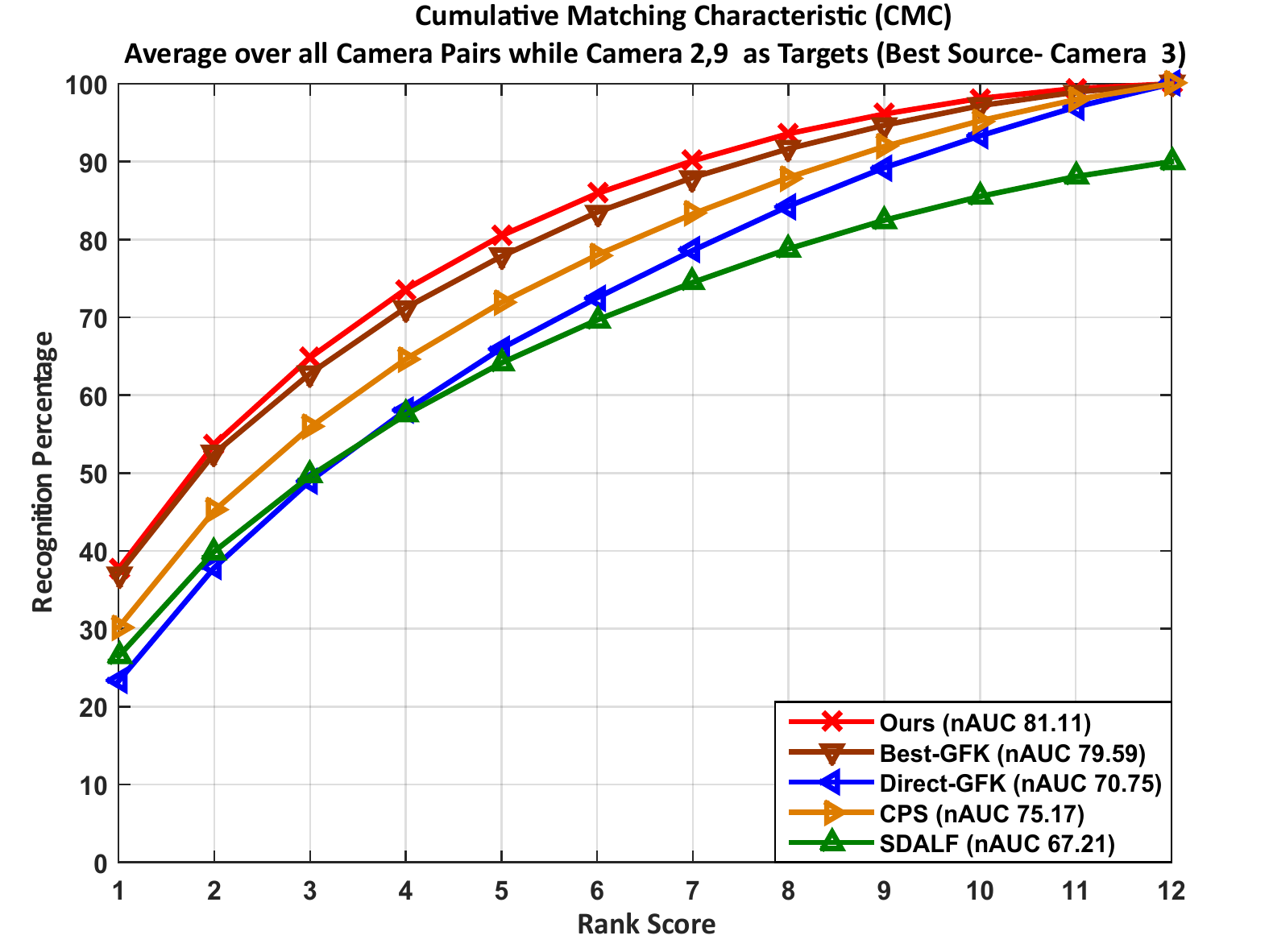} & 
			\includegraphics[height=.23\linewidth,width=.31\linewidth]{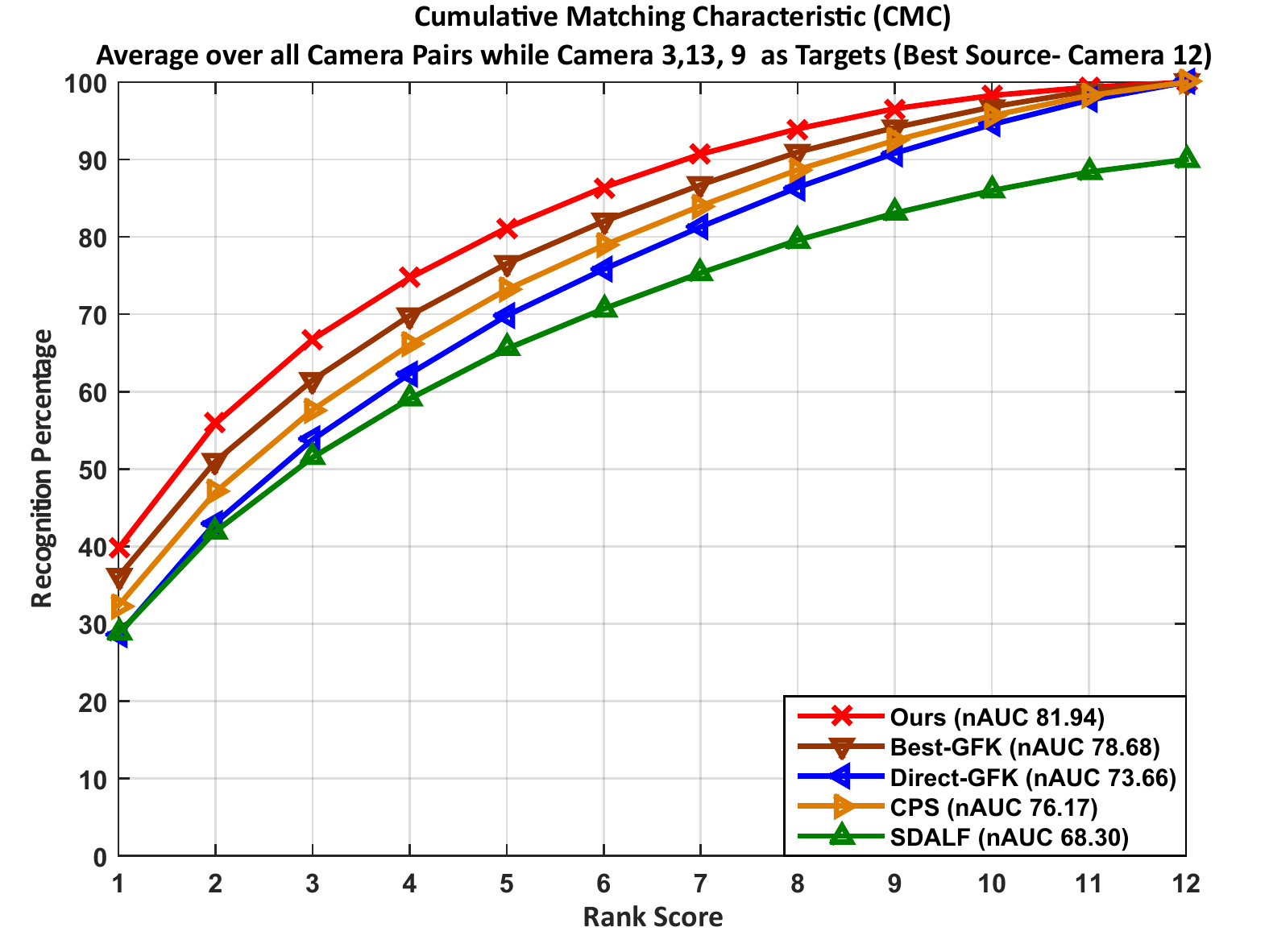}&
			\includegraphics[height=.23\linewidth,width=.31\linewidth]{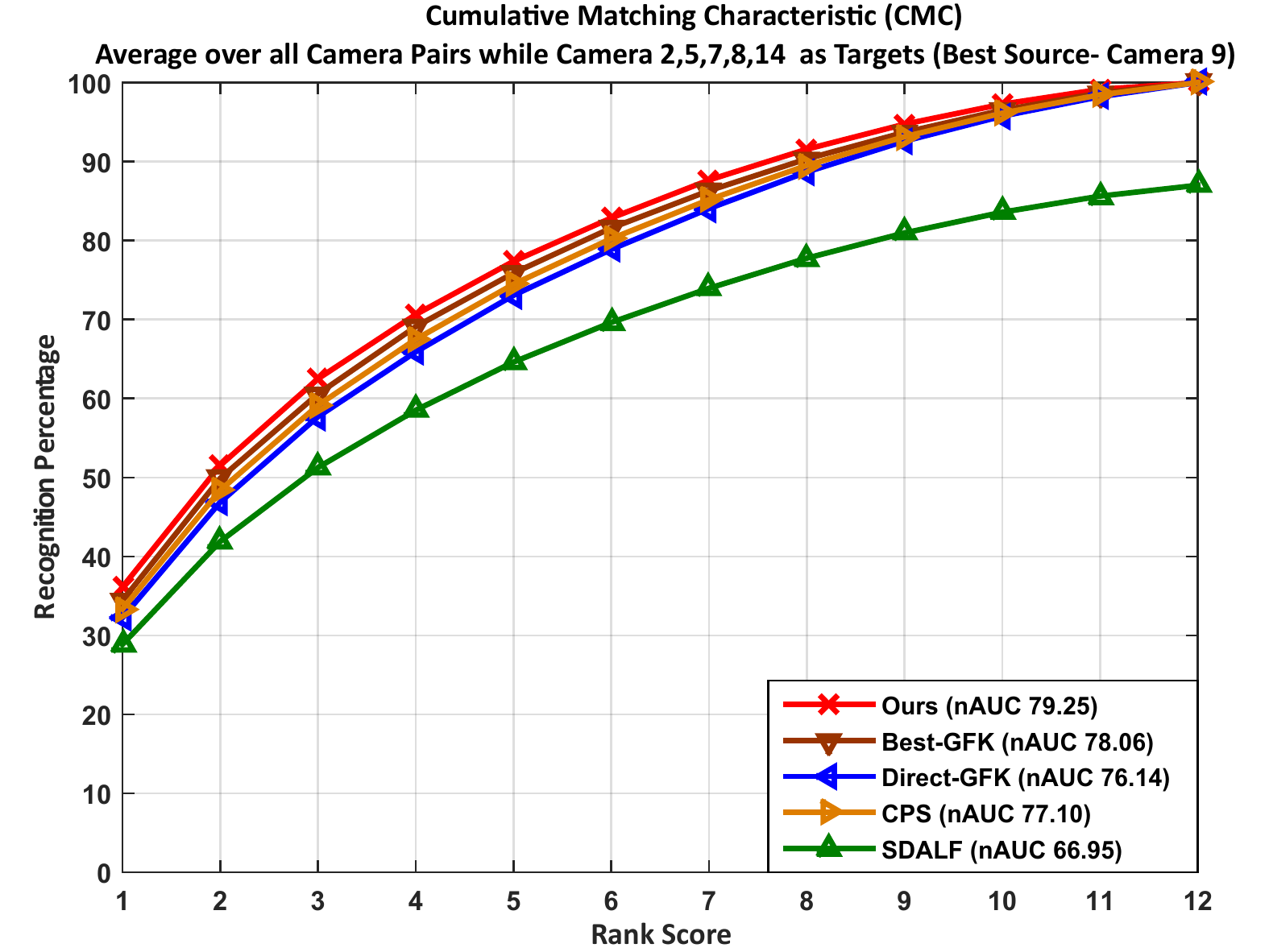}\\ 
			{\scriptsize (a) Camera 2, 9 as Targets } & \scriptsize{ (b) Camera 3, 9, 13 as Targets}  & \scriptsize{ (c) Camera 2, 5, 7, 8, 14 as Targets}  \\
		\end{tabular}
	\end{center}
	\vspace{-2mm}
	\caption{CMC curves for Shinpuhkan2014 dataset with 16 cameras. Plots (a, b, c) show the performance of different methods while introducing 2, 3 and 5 cameras respectively at the same time. Please see the text in Section~\ref{sec:multi} for the analysis of the results.}
	\label{fig:Shinpuhkan}  \vspace{-4mm}
\end{figure*}

\textbf{Results.} Fig.~\ref{fig:WARD} shows the results for all possible 3 combinations (two source and one target) on the 3 camera WARD dataset, whereas Fig.~\ref{fig:RAiD-SAIVT} shows the average performance over all possible combinations by inserting one camera, on RAiD and SAIVT-SoftBio dataset, respectively. From all three figures, the following observations can be made: (i) the proposed framework for re-identification consistently outperforms all compared unsupervised methods on all three datasets by a significant margin. (ii) among the alternatives, \texttt{CPS} baseline is the most competitive. However, the gap is still significant due to the two introduced components working in concert: discovering the best source camera and exploiting its information for re-identification. The rank-1 performance improvements over \texttt{CPS} are 23.44\%, 24.50\% and 9.98\% on WARD, RAiD and SAVIT-SoftBio datasets respectively. (iii) \texttt{Best-GFK} works better than \texttt{Direct-GFK} in most cases, which suggests that kernels computed across the best source camera and target camera can be applied to find the matching accuracy across other camera pairs in re-identification. (iv) Finally, the performance gap between our method and \texttt{Best-GFK} (maximum improvement of 17\% in nAUC on RAiD) shows that the proposed transitive algorithm is effective in exploiting information from the best source camera while computing re-id accuracy across camera pairs. 

\vspace{-1mm}
\subsection{Introducing Multiple Cameras}
\label{sec:multi}

\textbf{Goal.} The aim of this experiment is to validate the effectiveness of our proposed approach while introducing multiple cameras at the same time in a dynamic camera network. 

\textbf{Implementation Details.} We conduct this experiment on Shinpuhkan2014 dataset~\cite{kawanishi2014shinpuhkan2014} with 16 cameras. We randomly chose 2, 3 and 5 cameras as the target cameras while remaining cameras are possible source cameras. For each case, we pick the common best source camera based on the average distance and follow the same strategy as in Section~\ref{sec:one}. Results with multiple best source cameras, one for each target camera, are included in the Supplementary. 

\textbf{Results.} Fig.~\ref{fig:Shinpuhkan} shows results of our method while randomly introducing 2, 3 and 5 cameras respectively on Shinpuhkan2014 dataset. From Fig.~\ref{fig:Shinpuhkan} (a, b, and c), the following observations can be made: (i) Similar to the results in Section~\ref{sec:one}, our approach outperforms all compared methods in all three scenarios. This indicates that the proposed method is very effective and can be applied to large-scale dynamic camera networks where multiple cameras can be introduced at the same time. (ii) The gap between ours and \texttt{Best-GFK} is moderate but still we improve by 4\% in nAUC values, which corroborates the effectiveness of transitive inference for re-identification in a large-scale camera network.   
\vspace{-1mm}
\subsection{Extension to Semi-supervised Adaptation}
\label{sec:semiexp}

\textbf{Goal.} The proposed method can be easily extended to semi-supervised settings when labeled data from the target camera become available. The objective of this experiment is to analyze the performance of our approach in such settings by incorporating labeled data from the target domain.

\textbf{Compared Methods.} We compare the proposed unsupervised approach with four variants of our method  where 10\%, 25\%, 50\% and 100\% of the labeled data from target camera are used for estimating kernel matrix respectively. 

\textbf{Implementation Details.} We follow same experimental strategy in finding average re-id accuracies over a camera network. However, we use PLS instead of PCA, to compute the discriminative subspaces in target camera by considering 10\%, 25\%, 50\% and 100\% labeled data respectively.

\begin{figure}
	\begin{center}
		\begin{tabular}{cc}
			\includegraphics[scale=.245]{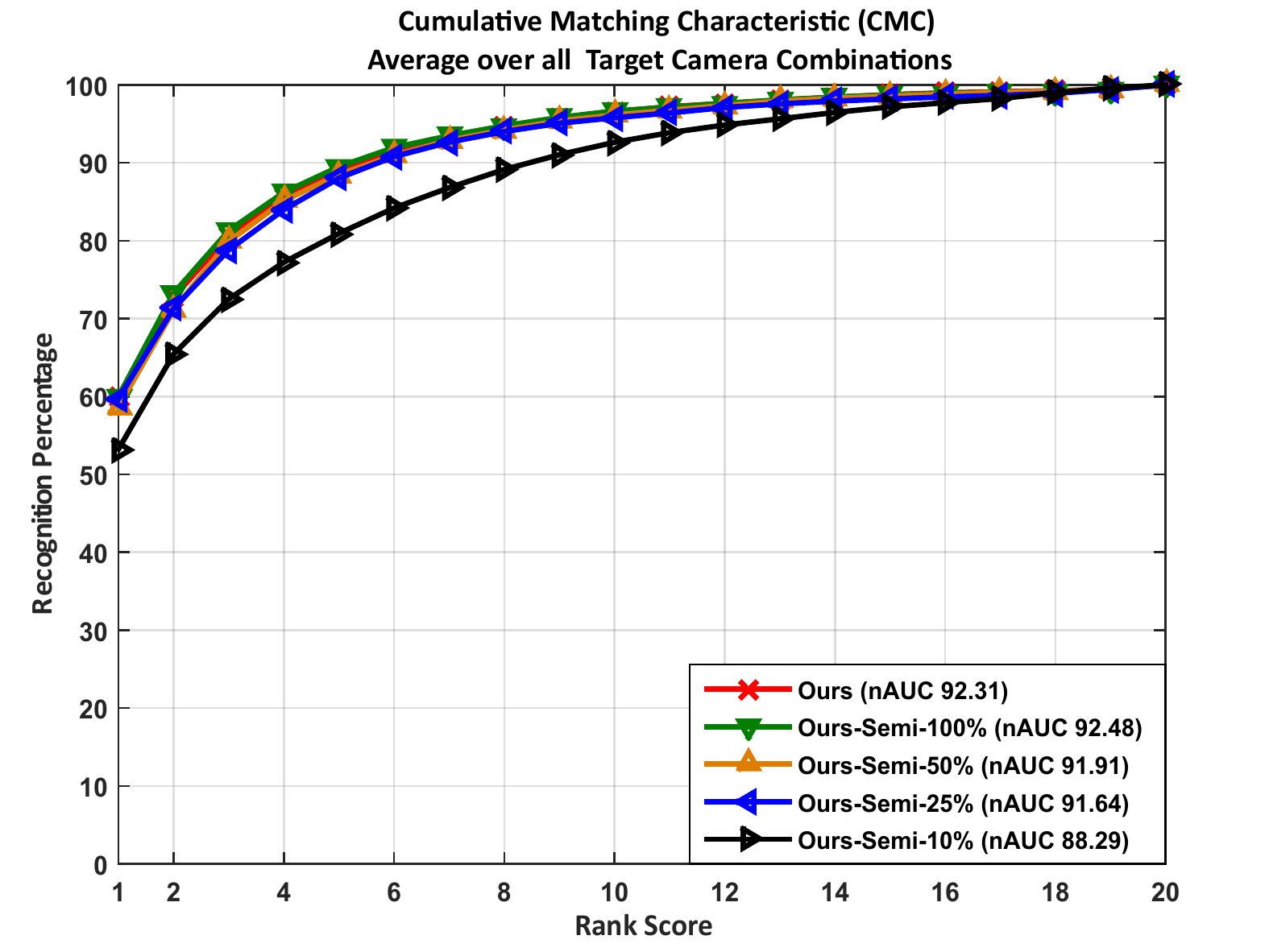} & 
			\includegraphics[scale=.245]{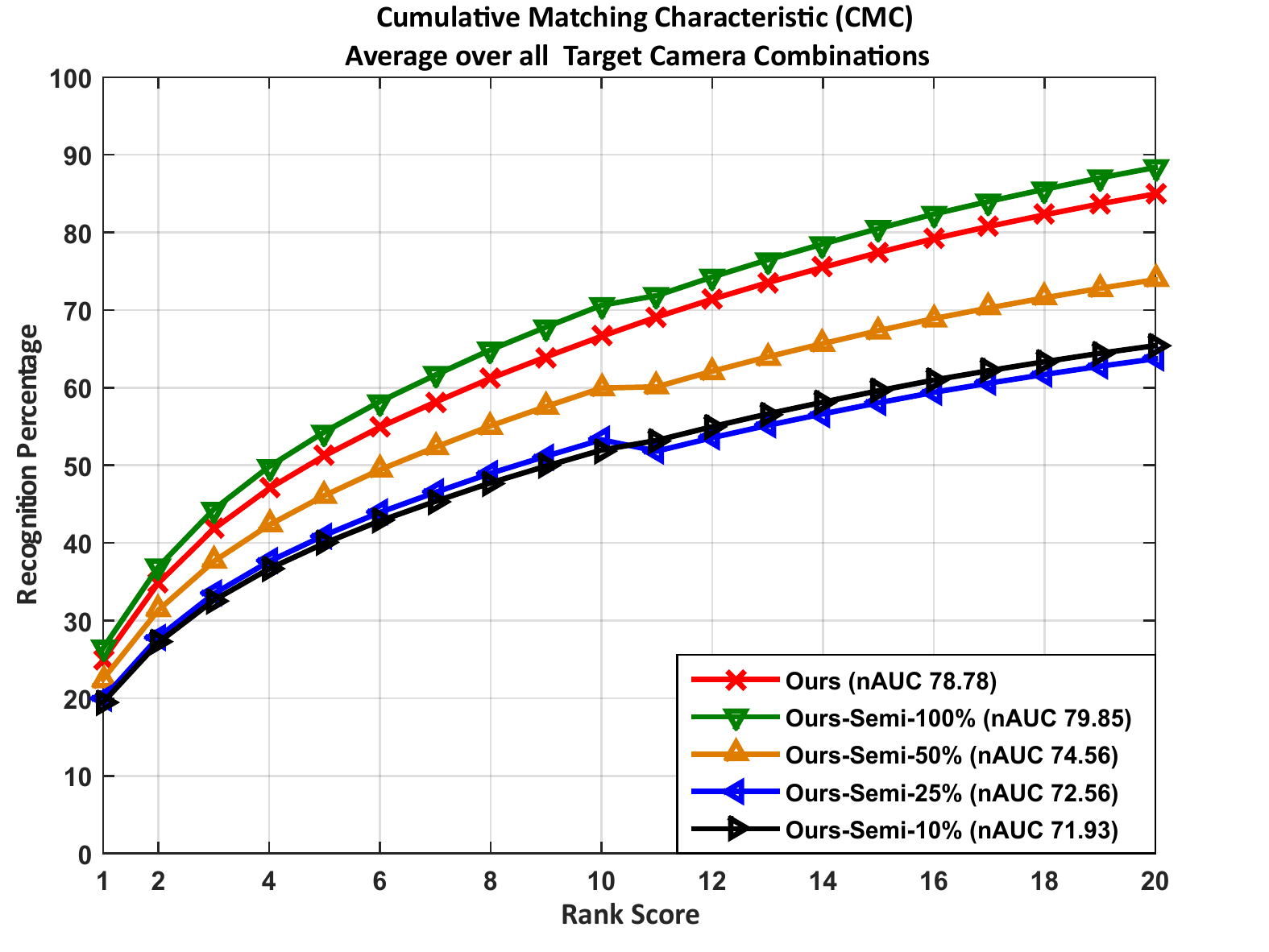} \\
			{\scriptsize (a) RAiD } & \scriptsize{ (b) SAIVT-SoftBio} \\
		\end{tabular}
	\end{center}
	\vspace{-3mm}
	\caption{Semi-supervised adaptation with labeled data. Plots (a,b) show CMC curves averaged over all target camera combinations, introduced one at a time, on RAiD and SAVIT-SoftBio respectively. Please see the text in Section~\ref{sec:semiexp} for analysis of the results.}
	\label{fig:Avg-Semi}  \vspace{-3mm}
\end{figure}

\textbf{Results.} We have the following key findings from Fig.~\ref{fig:Avg-Semi}: (i) As expected, the semi-supervised baseline \texttt{Ours-Semi-100\%}, works best since it uses all the labeled data from target domain to compute the kernel matrix for finding the best source camera. (ii) Our method remains competitive to \texttt{Ours-Semi-100\%} on both datasets (Rank-1 accuracy: 60.04\% vs 59.84\% on RAiD and 26.41\% vs 24.92\% on SAVIT-SoftBio). 
However, it is important to note that collecting labeled samples from the target camera is very difficult in practice.
(iii) Interestingly, the performance gap between our unsupervised method and other three semi-supervised baselines (\texttt{Ours-Semi-50\%}, \texttt{Ours-Semi-25\%}, and \texttt{Ours-Semi-10\%}) are moderate on RAiD (Fig.~\ref{fig:Avg-Semi}-a), but on SAVIT-SoftBio, the gap is significant (Fig.~\ref{fig:Avg-Semi}-b). We believe this is probably due to the lack of enough labeled data in the target camera to give a reliable estimate of PLS subspaces.

\vspace{-1mm}
\subsection{Re-identification with LDML Metric Learning}
\label{sec:lfda} 
\textbf{Goal.} The objective of this experiment is to verify the effectiveness of our approach by changing the initial setup presented in Section~\ref{sec:Preliminaries}. Specifically, our goal is to show the performance of the proposed method by replacing KISSME~\cite{kostinger2012large} with LDML metric learning~\cite{guillaumin2009you}. Ideally, we would expect similar performance improvement by our method, irrespective of the metric learning used to learn the optimal distance metrics in an existing network of cameras. 

\begin{figure}
	\begin{tabular}{cc}
		\includegraphics[scale=.245]{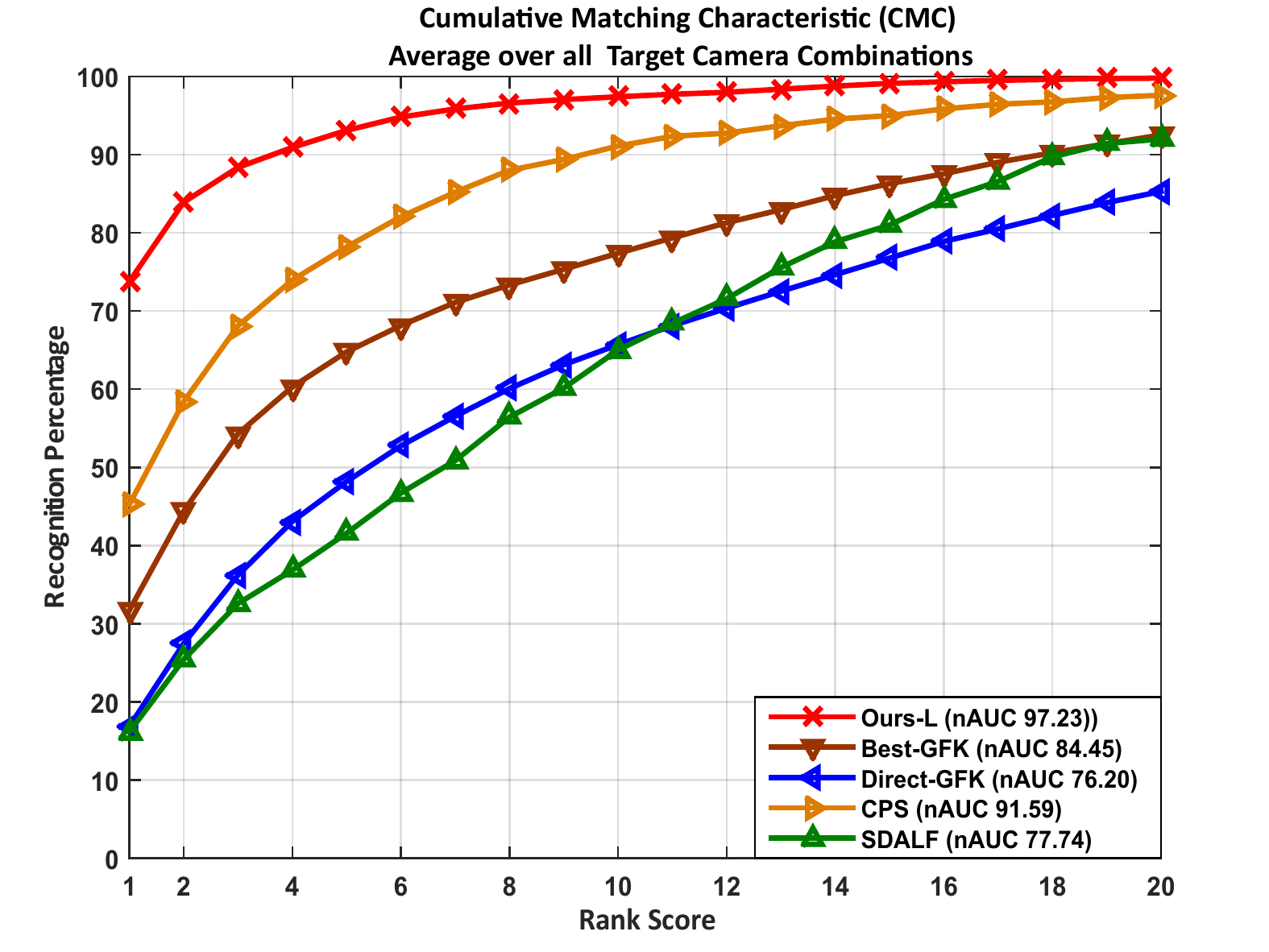}& 
		\includegraphics[scale=.245]{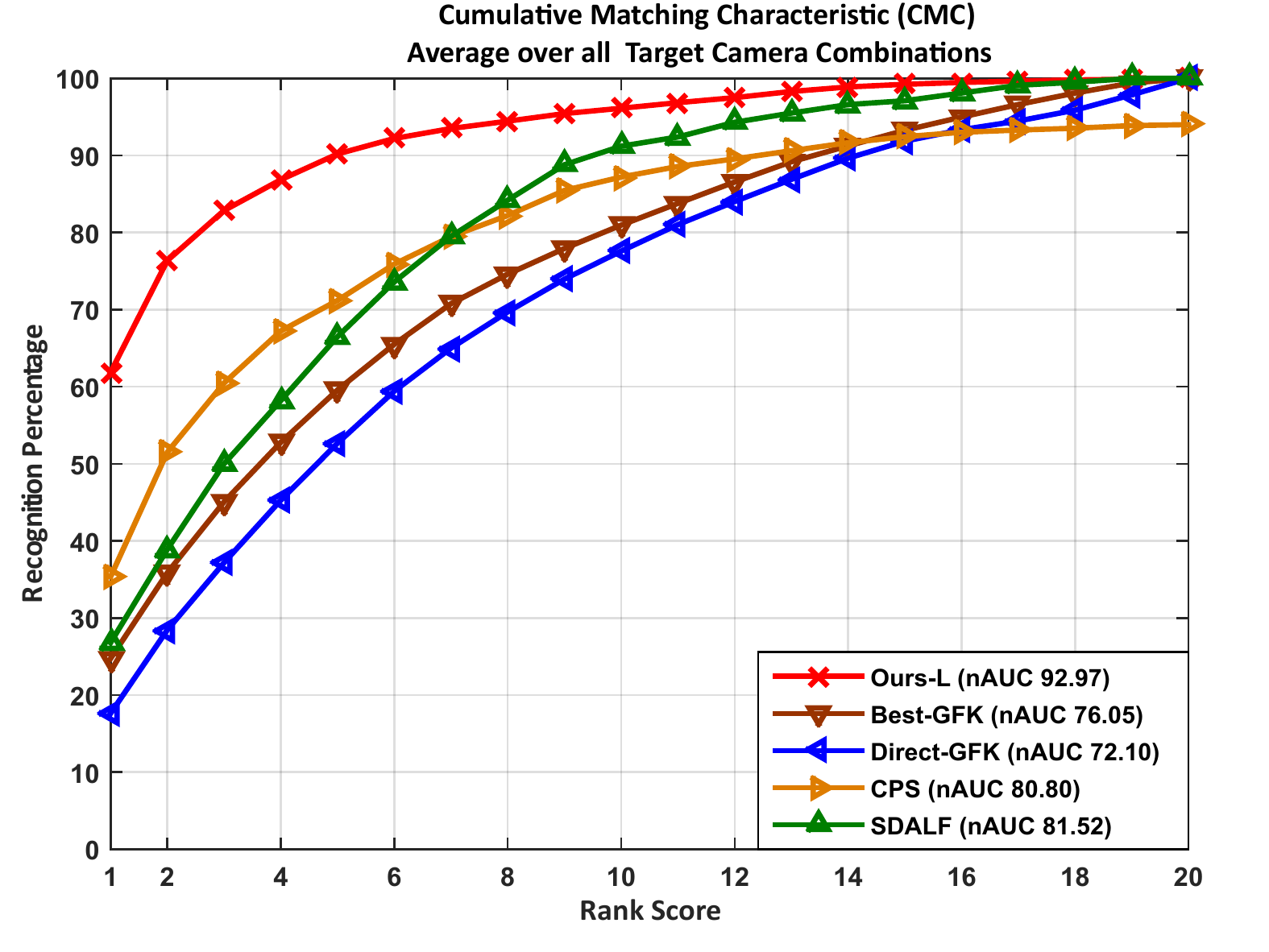}\\ 
		{\scriptsize (a) WARD } & \scriptsize{ (b) RAiD}   \\
	\end{tabular}
	\vspace{-2mm}
	\caption{Re-id performance with LDML as initial setup. Plots (a,b) show CMC curves averaged over all target camera combinations, introduced one at a time, on WARD and RAiD respectively. 
		}
	\label{fig:Avg-LDML}  \vspace{-4mm}
\end{figure}
\textbf{Results.} Fig.~\ref{fig:Avg-LDML} shows results on WARD and RAiD respectively. Following are the analysis of the figures: (i) Our approach outperforms all compared methods in both datasets which suggests that the proposed adaptation technique works significantly well irrespective of the metric learning method used in the existing camera network. (ii) The proposed approach works slightly better with LDML compared to KISSME on WARD dataset (73.77 vs 68.99 in rank-1 accuracy). However, the margin becomes smaller on RAiD (61.87 vs 59.84) which is relatively a complex re-id dataset with 2 outdoor and 2 indoor cameras. (iii) Although performance of LDML is slightly better than KISSME, it is important to note that KISSME is about 40\% faster than that of LDML in learning the metrics in WARD dataset. 
KISSME is computationally efficient and hence more suitable for learning metrics in a large-scale camera network.      
\vspace{-2mm}
\subsection{Comparison with Supervised Re-identification}
\label{sec:sup}      
\textbf{Goal.} The objective of this experiment is to compare the performance of our approach with supervised alternatives in a dynamic camera network.

\textbf{Compared Methods.} We compare with several supervised alternatives which fall into two categories: (i) feature transformation based methods including \texttt{FT}~\cite{martinel2015re}, \texttt{ICT}~\cite{avraham2012learning}, \texttt{WACN}~\cite{6239203}, that learn the way features get transformed between two cameras and then use it for matching, (ii) metric learning based methods including \texttt{KISSME}~\cite{kostinger2012large}, \texttt{LDML}~\cite{guillaumin2009you}, \texttt{XQDA}~\cite{lisanti2015person} and \texttt{MLAPG}~\cite{liao2015efficient}. As mentioned earlier, our model can operate with any initial network setup and hence we show our results with both KISSME and LDML, denoted as \texttt{Ours-K} and \texttt{Ours-L}, respectively. Note that we could not compare with recent deep learning based methods as they are mostly specific to a static setting and also their pairwise camera results are not available on the experimented datasets. We did not re-implement such methods in our dynamic setting as it is very difficult to exactly emulate all the implementation details.     

\textbf{Implementation Details.} To report existing feature transformation based methods results, we use prior published performances from~\cite{das2014consistent}. For metric learning based methods, we use publicly available codes and test on our experimented datasets. Given a newly introduced camera, we use the metric learning based methods to relearn the pair-wise distance metrics using the same train/test split, as mentioned earlier in experimental settings. For each datasets, we show the average performance over all possible combinations by introducing one camera at a time.
\begin{table}[t]
	\centering
	\scriptsize
	\caption{Comparison with supervised methods. Numbers show rank-1 recognition scores in \% averaged over all possible combinations of target cameras, introduced one at a time. 
	}
	\label{tab:Rank1}
	\begin{tabulary}{1.1\linewidth}{|p{11.9mm}||P{10mm}|P{10mm}|P{20.5mm}|}
		\hline
		\textbf{Methods} &\textbf{WARD} & \textbf{RAiD} &\textbf{Reference}\\
		\hhline{|=|=|=|=|}
		\texttt{FT}		& 49.33	& 39.81	&TPAMI2015~\cite{martinel2015re}\\
		\texttt{ICT}	& 42.51	& 25.31 &ECCV2012~\cite{avraham2012learning}\\
		\texttt{WACN}	& 37.53	& 17.71	& CVPRW2012~\cite{6239203}  \\
		\hhline{|-|-|-|-|}
		\texttt{KISSME}	& 66.95	& 55.68	& CVPR2012~\cite{kostinger2012large} \\
		\texttt{LDML}	& 58.66	& 61.52	& ICCV2009~\cite{guillaumin2009you} \\
		\texttt{XQDA}	& \textbf{77.20}	& \textbf{77.81}	& TPAMI2015~\cite{lisanti2015person} \\
		\texttt{MLAPG}	& 72.26	& 77.68	& ICCV2015~\cite{liao2015efficient} \\
		\hhline{|=|=|=|=|}
		\texttt{Ours-K} & \textit{68.99} & \textit{59.84} &Proposed\\
		\texttt{Ours-L} & \textit{73.77} & \textit{61.87} &Proposed\\
		\hline 		
	\end{tabulary}  
	\vspace{-4mm}
\end{table}

\textbf{Results.} Table~\ref{tab:Rank1} shows the rank-1 accuracy averaged over all possible target cameras introduced one at a time in a dynamic network. We have the following key findings from Table~\ref{tab:Rank1}: (i) Both variants of our unsupervised approach (\texttt{Ours-K} and \texttt{Ours-L}) outperforms all the feature transformation based approaches on both datasets by a big margin. (ii) On WARD dataset with 3 cameras, our approach is very competitive on both settings: \texttt{Ours-K} outperforms \texttt{KISSME} and \texttt{LDML} whereas \texttt{Ours-L} overcomes \texttt{MLAPG}. This result suggests that our approach is more effective in matching persons across a newly introduced camera and existing source cameras by exploiting information from best source camera via a transitive inference. (iii) On the RAiD dataset with 4 cameras, the performance gap between our method and metric-learning based methods begins to increase. This is expected as with a large network involving a higher number of camera pairs, an unsupervised approach can not compete with a supervised one, especially, when the latter one is using an intensive training phase. However, we would like to point out once more that in practice collecting labeled samples from a newly inserted camera is very difficult and unrealistic in actual scenarios. 

We refer the reader to the supplementary material
for more detailed results (individual CMC curves) along with qualitative matching results on all datasets.

\vspace{-1mm}
\section{Conclusions}
\label{sec:Conclusion}
\vspace{-1mm}
We presented an unsupervised framework to adapt re-identification models in a dynamic network, where a new camera may be temporarily inserted into an existing system to get additional information. We developed a domain perceptive re-identification method based on geodesic flow kernel to find the best source camera
to pair with a newly introduced target camera, without requiring a very expensive training phase. In addition, we introduced a simple yet effective transitive inference algorithm that can exploit information from best source camera to improve the accuracy across other camera pairs. Extensive experiments on several benchmark datasets well demonstrate the efficacy of our method over state-of-the-art methods. 

\vspace{1mm}
\textbf{Acknowledgements:} This work is partially supported by NSF grant CPS 1544969. 
\newpage

{\small
\bibliographystyle{ieee}
\bibliography{egbib_amran}
}

\end{document}